\def\year{2019}\relax
\documentclass[letterpaper]{article} 
\usepackage{aaai19}  
\usepackage{times}  
\usepackage{helvet}  
\usepackage{courier}  
\usepackage{url}  
\usepackage{graphicx} 
\usepackage{breqn}
\usepackage{amssymb}
\title{Low Precision Policy Distillation with Application to Low-Power, Real-time Sensation-Cognition-Action Loop with Neuromorphic Computing}

\author{Jeffrey L. Mckinstry\\ IBM Research\\ jkmckins@us.ibm.com \\\\{\bf \Large Steven K. Esser} \\ IBM Research \\ sesser@us.ibm.com
  \And Davis R. Barch \\ IBM Research \\ drbarch@us.ibm.com \\\\{\bf \Large Jeffrey A. Kusnitz} \\ IBM Research \\ jk@us.ibm.com
  \And Deepika Bablani \\ IBM Research \\ deepika.bablani@ibm.com \\\\{\bf \Large John V. Arthur} \\ IBM Research \\ arthurjo@us.ibm.com
 \And Michael V. Debole \\IBM Research \\ mvdebole@us.ibm.com \\\\{\bf \Large Dharmendra S. Modha} \\ IBM Research \\ dmodha@us.ibm.com
 }
 
\begin{document}
 \maketitle
\begin{abstract}
  Low precision networks in the reinforcement learning (RL) setting are relatively unexplored because of the limitations of binary activations for function approximation.  Here, in the discrete action ATARI domain, we demonstrate, for the first time, that low precision policy distillation from a high precision network provides a principled, practical way to train an RL agent.  As an application, on 10 different ATARI games, we demonstrate real-time end-to-end game playing on low-power neuromorphic hardware by converting a sequence of game frames into discrete actions.
\end{abstract}

\section{Introduction}
In recent years, deep learning has contributed to unprecedented success in a wide variety of machine learning applications including computer vision, natural language processing, recommender systems and speech. Another area of growing interest where contributions of deep learning have led to noteworthy results is that of sequential decision making through deep reinforcement learning and control. One of the early notable examples of this is training Deep Q Networks (DQN) to play ATARI games \cite{mnih2015human}. In order for these advances to be successfully applied towards solving real world problems, it is important to be able to deploy these algorithms on hardware that allows for high performance in real time while remaining energy efficient. 

We use knowledge distillation to apply existing reinforcement learning solutions to neuromorphic hardware. Knowledge distillation has been used previously by \cite{rusu2015policy} in the reinforcement learning setting for training smaller student networks to learn a teacher network's policy as illustrated in Figure \ref{fig:policy_distillation}. Policy distillation is particularly attractive in the low precision reinforcement learning setting because it reduces the Q value regression problem to a supervised learning problem of imitating a teacher’s policy, which is simpler. It is therefore possible to leverage proven low-precision network training algorithms \cite{courbariaux2015binaryconnect},\cite{rastegari2016xnor}, which successfully train networks with binary activations, ternary weights, and fixed fan-in, ideal for low-power hardware implementation \cite{esser2015backpropagation}.

\begin{figure}[ht]
  \centering
 \includegraphics[width=0.4\textwidth, ,height=0.3\textwidth]{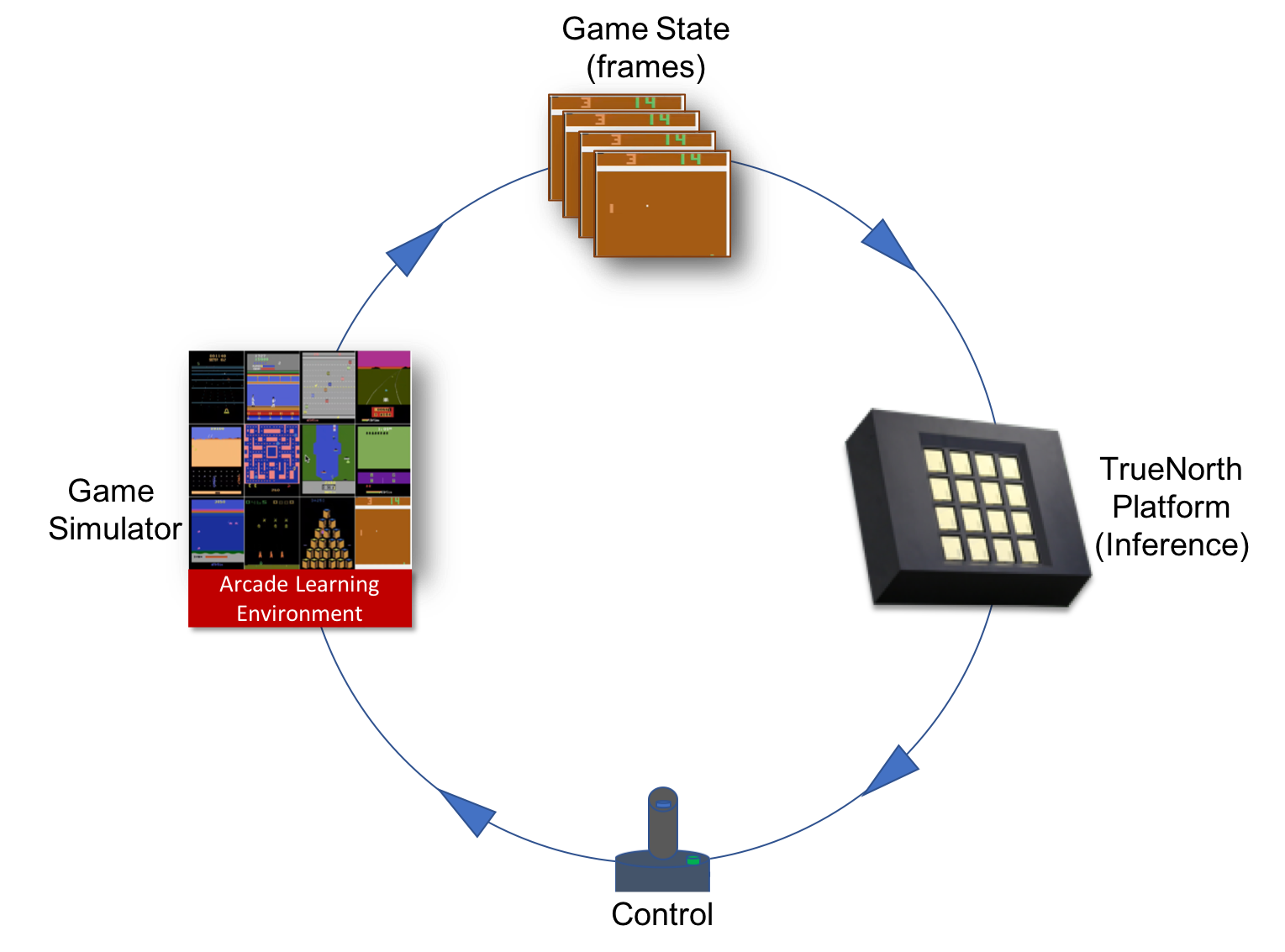}
  \caption{\textbf{Real-time Sensation-Cognition-Action loop with TrueNorth allows real world deployment of RL models.} The TrueNorth system receives as input the game state from the game simulator (sensation), processes the frames to generate predicted return from all possible moves in the game (cognition) and outputs the best move for that state based on its learned policy (action) in real time.}
  \label{fig:hardware}
\end{figure}

\begin{figure*}[ht]
  \centering
 \includegraphics[width=0.75\textwidth, ,height=0.45\textwidth]{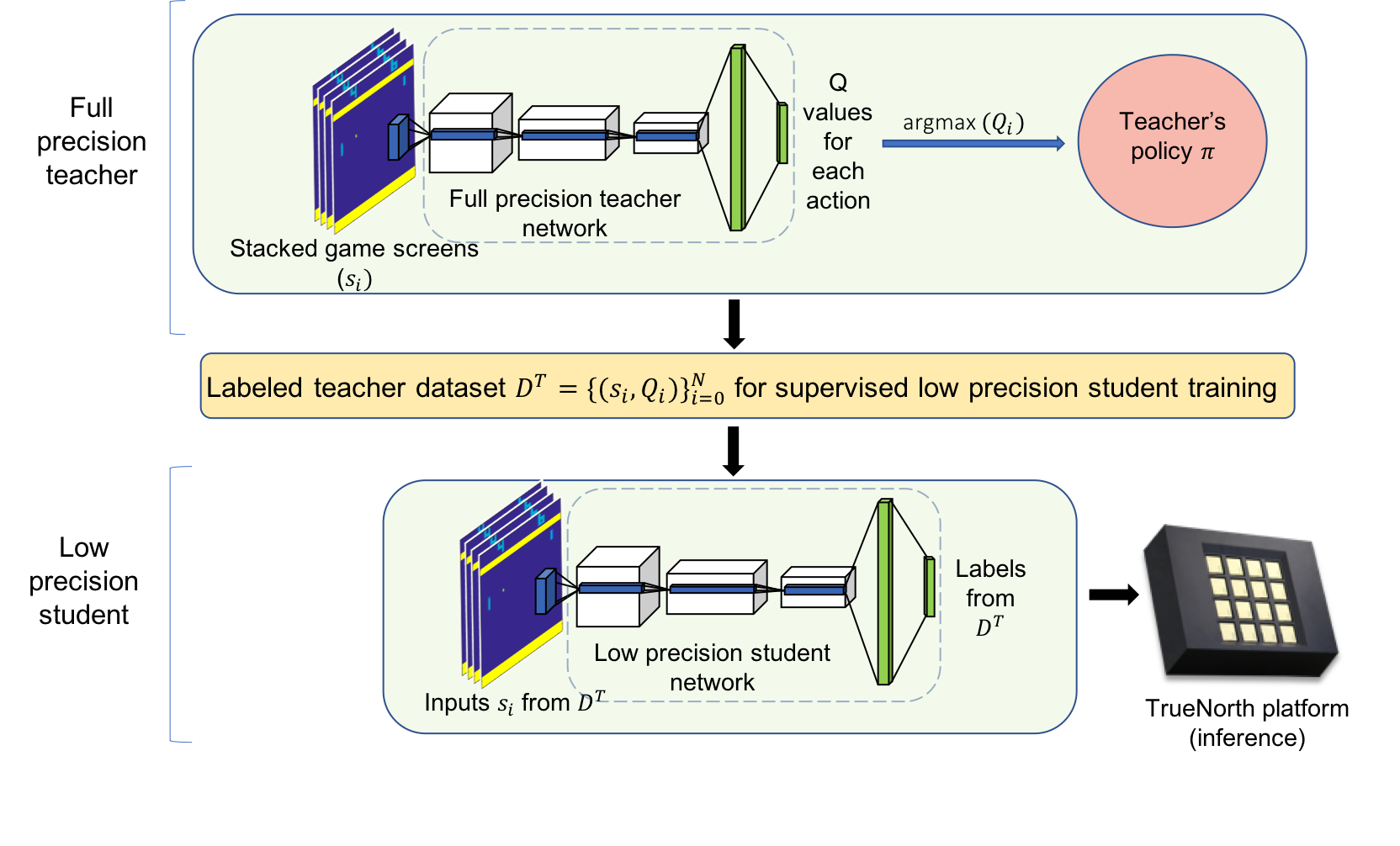}
  \caption{\textbf{Low precision policy distillation allows principled training of constrained networks for neuromorphic systems.} (Top) Full precision teacher network trained using value based RL (Double DQN in this work). (Bottom) Low precision student network, once trained on the teacher's policy, is mapped to the TrueNorth platform.}
  \label{fig:policy_distillation}
\end{figure*}

\subsection{Contributions}
We test the proposed solution in the ATARI games environment.  We find
\begin{itemize}
\item Policy distillation produces solutions which are within two percent of the teacher's score on average, and in some cases surpass the teacher's score
\item The scores obtained by the low-precision student networks are a function of the network capacity
\item Similar to \cite{rusu2015policy} we find that for low-precision training, the KL-divergence loss gives superior results to Negative Log Likelihood loss
\item A single, capacity-limited, low-precision network can learn to play multiple games with the distillation approach
\item By using policy distillation, we were able to create an end-to-end, real-time demonstration system with neuromorphic hardware playing ATARI games (Figure \ref{fig:hardware})
\end{itemize}
To summarize, we sidestep the potential difficulties of directly training binary neural network as function approximators (see background section and \cite{blum1991approximation}) by using policy distillation. It provides a principled approach to train low-precision networks, and we utilize this approach to demonstrate for the first time an end to end system for playing ATARI games on energy efficient hardware, running in real time.

It is important to note that by employing distillation and effectively employing the teacher only to generate the training data, we make our approach agnostic to the choice of reinforcement learning algorithm used to train the teacher. As a result, even though we present experimental results only for DQN, we can essentially incorporate any state of the art reinforcement learning algorithm to train the teacher and achieve improved student results, eliminating the need to individually  adapt every algorithm to the low precision setting.

\section{Background}

\subsection{Value based RL is inefficient with binary neural networks}
Low precision Convolutional Neural Networks (CNNs) have recently been shown to achieve near state-of-the-art on a wide range of machine-learning benchmarks \cite{courbariaux2015binaryconnect},\cite{rastegari2016xnor}, despite binary activations, ternary weights, and fixed fan-in, ideal for low-power hardware implementation \cite{esser2015backpropagation}. The ability to apply such low precision networks to value-based RL algorithms \cite{mnih2013playing}, \cite{van2016deep} remains unclear for the following reasons.  
\begin{itemize}
  \item Such algorithms use CNNs to approximate the action value, $Q(s,a)$ for each state and action, which is the estimated future reward obtained by executing action $a$ given the current state, $s$.  These action values are continuous, with ranges that are problem-dependent, making value-based RL a challenging regression task which is inherently quite different from the classification tasks explored in the low precision literature. 

  \item In \cite{blum1991approximation}, the authors prove that multi-layer neural networks with binary activations perform piecewise constant approximation and are universal function approximators, indicating that they can theoretically be used for regression. It is also known that the error for piecewise constant approximation decreases only linearly as a function of the number of intervals as opposed to quadratically in the case of piece-wise linear function approximation\cite{quarteroni2010numerical}. This means that to obtain the same accuracy as a network with $n$ neurons with ReLU activation functions, a network with binary activations would require on the order of $n^2$ neurons, suggesting that using binary neurons for Q function approximation can be inefficient.
  
  \item Finding a solution for the sequential decision making problem in this constrained space is inherently hard owing to the non-stationary data distribution, limited feedback and delayed rewards \cite{mnih2013playing}, and assuming back-propagation can be used to find a solution in this space, the amount of time required by back-propagation can be prohibitively high. 
\end{itemize}

\subsection{Deep Q Networks and Double Deep Q Networks}
RL solves a sequential decision making task, in which an agent interacts with an environment $E$ over discrete time steps \cite{sutton1998reinforcement}. For this work, we consider the ATARI domain, which is a well studied benchmark in deep reinforcement learning literature, including \cite{mnih2013playing}, \cite{van2016deep}; \cite{bellemare2016increasing}, \cite{schaul2015prioritized}. In this setting, at every time step, the RL agent receives from the environment an image frame of the game screen as raw input. The agent then uses this frame $x_t$ at time $t$, along with $N$ preceding frames $(x_{t-1},x_{t-2},...,x_{t-N})$ to choose the optimal action from a discrete set of $N$ actions $A = \{a_1, a_2, ... , a_N\}$ and receives a reward $r_t$ from the environment. The optimal action is chosen to maximize the expected long term reward $R_t$ from the environment. When an agent is acting according to a stochastic policy $\pi$, the value of taking an action $a$ in a state $s$ is defined by a Q value, while the value of the state is defined by V, as follows
\begin{equation}
    Q^\pi(s,a) = \mathbb{E}[R_t|s_t=s,a_t=a,\pi]
\end{equation}
\begin{equation}
    V^\pi(s) = \mathbb{E}_{a\sim\pi(s)}[Q^\pi(s,a)]
\end{equation}

Deep regression networks have been used successfully to approximate $Q$ functions, introduced first in \cite{mnih2013playing}. These networks, parameterized by $\theta$, are convolutional neural networks that take raw game frames ($s_t$) as input and predicted the $Q$ value for each action in the input state, which is then used to choose the optimal action by balancing exploration and exploitation. To break the correlation between sequential input data and stabilize training, \cite{mnih2013playing} store samples of $(s,a,r,s')$ in an  \textit{experience replay} memory buffer and train the network from samples drawn from the replay memory using the loss function below.

\begin{dmath}
    L_i(\theta_i) = \mathbb{E}_{(s,a,r,s')\sim U(D)} \left[ \left( r + \gamma \textrm{max}_{a'}Q(s',a';\theta_{i}^-) 
    - Q(s,a;\theta_{i}) \right)^2 \right]
\end{dmath}

Here $\gamma$ is the discount factor $R_t = \sum_{t'=t}^T$ $\gamma^{t'-t}$ $r_t$ and $\theta_{i}^-$ are the older parameters of a frozen \textit{target network}. Using these ingredients, DQN achieved super human performance in many games in the ATARI suite. For our work, we use the Double DQN (DDQN) learning algorithm introduced in \cite{van2016deep}, an improvement over DQN that prevents overoptimistic Q estimates of DQN by using different networks to select and evaluate an action. 

\subsection{Low Precision Networks for Energy Efficient Hardware}
Interest in low precision hardware for energy efficient deployment of deep learning algorithms in real world applications has been on the rise in the past few years. Recent work \cite{courbariaux2015binaryconnect}, \cite{rastegari2016xnor}, \cite{esser2015backpropagation}  has demonstrated that low precision networks can be trained on standard machine learning tasks. 

In \cite{esser2015backpropagation} the authors show that backpropagation can be adapted to train low precision, deep convolutional networks, achieving near state-of-the-art performance in a wide variety of tasks using the TrueNorth chip \cite{merolla2014million}. TrueNorth is a non-von Neumman architecture, inspired by biological brains and designed for deploying low precision event-driven neural networks that operate at very low power. Like the brain, the architecture co-localizes memory, compute, and communication within its fundamental unit, a neurosynaptic core. Each TrueNorth chip integrates 4,096 parallel cores interconnected  with an on-chip network, providing a flexible platform that supports neural networks with  over 1 million programmable spiking neurons and over 256 million configurable synapses. While processing 1000
$32 \times 32$ pixel, three color images per second, a TrueNorth chip operates with a total power consumption on the order of 100mW\cite{Esser11441}. 

As a result of its hardware constraints, the networks deployed on the chip must satisfy the constraints imposed by the architecture- neurons output binary events, synapses have low precision, connectivity is core-to-core, not all-to-all. In the context of this work, \textit{low precision} refers to this constrained parameter setting, i.e., binary activations, ternary weights (and limited input-output connectivity). Convolutional neural networks are mapped to TrueNorth using the Energy efficient deep networks (EEDN) algorithm \cite{Esser11441}. The EEDN algorithm satisfies the TrueNorth architecture constraints by restricting network precision to binary neuron output and ternary weights $\{−1, 0, 1\}$, and by limiting neuron fan-in and fan-out.

\begin{itemize}
    \item \textbf{Binary output} : Instead of the standard rectified linear unit (ReLU) activation function, EEDN uses a binary step function implemented as threshold logic units with integer biases. The derivative for backpropagation, delta function, is approximated with a triangular function $\frac{dy}{dr} = \max(0, 1)$ where $r$ is the filter response and $y$ is the output. Filter output is computed using batch normalization \cite{ioffe2015batch} during training, with the batch normalization parameters rolled into the neuron threshold for deployment.
    \item \textbf{Ternary weights} : Ternary weights $w$ are trained offline using an adaptation of the standard backpropagation algorithm. The ternary weights are used during the forward and backpropagation passes. However, the resulting weight updates are applied to a shadow network of high-precision proxy weights $w_h$.
    \item \textbf{Neuron fan-in} :  Group constraints are used to accommodate the limited fan-in constraint of 128 inputs per neuron (while TrueNorth allows 256 inputs per neuron, two inputs are used per synapse to allow ternary weights). Specifically, in a layer which uses $G$ groups, each neuron receives connections from only $N/G$ features from the $N$ features in the source layer. For example, a convolution layer with a $1 \times 1$ kernel receiving input from 256 features in the source layer must use 2 groups in order to reduce the fan-in to $1 \times 1 \times  256/2 = 128$ inputs per neuron
    \item \textbf{Neuron fan-out} : TrueNorth neurons can target a single core. Therefore neuron copies are used to provide multiple neuron outputs for the weight representation scheme, and where filter overlap necessitates targeting multiple cores. A neuron is copied by replicating its parameters using free neurons on its own core, or by using splitter neurons on additional cores.
\end{itemize}


While non-trivial to train, these networks are extremely energy efficient in deployment and can function in real time. In this work, we use TrueNorth chip to deploy our networks, although any neuromorphic or low precision hardware can be used instead.

\begin{table*}[ht]
\centering
\resizebox{\textwidth}{!}{
\begin{tabular}{|c|c|c|c|c|c|c|c|c|c|c|c|c|c|c|c|c|c|c|c|}
\hline
\multicolumn{5}{|c|}{\textbf{1 chip}}                                                                                                                                                                                                                                     & \multicolumn{5}{c|}{\textbf{2 chip}}                                                                                                                                                                                                                                                     & \multicolumn{5}{c|}{\textbf{4 chip}}                                                                                                                                                                                                                                                                    & \multicolumn{5}{c|}{\textbf{8 chip}}                                                                                                                                                                                                                                                                    \\ \hline
\begin{tabular}[c]{@{}c@{}}Number\\ of\\ Features\end{tabular} & \begin{tabular}[c]{@{}c@{}}Kernel \\ Size\end{tabular} & Stride                                        & Pad                                           & Groups                                          & \begin{tabular}[c]{@{}c@{}}Number \\ of \\ Features\end{tabular} & \begin{tabular}[c]{@{}c@{}}Kernel \\ Size\end{tabular} & Stride                                            & Pad                                               & Groups                                               & \begin{tabular}[c]{@{}c@{}}Number\\  of \\ Features\end{tabular}  & \begin{tabular}[c]{@{}c@{}}Kernel \\ Size\end{tabular} & Stride                                                & Pad                                                   & Groups                                                     & \begin{tabular}[c]{@{}c@{}}Number \\ of \\ Features\end{tabular}  & \begin{tabular}[c]{@{}c@{}}Kernel \\ Size\end{tabular} & Stride                                                & Pad                                                   & Groups                                                     \\ \hline
32                                                             & 8                                                      & 4                                             & 2                                             & 1                                               & 32                                                               & 8                                                      & 4                                                 & 2                                                 & 1                                                    & 128                                                               & 8                                                      & 2                                                     & 1                                                     & 1                                                          & 256                                                               & 8                                                      & 2                                                     & 1                                                     & 1                                                          \\ \hline
\begin{tabular}[c]{@{}c@{}}256\\ 256\end{tabular}              & \begin{tabular}[c]{@{}c@{}}4\\ 1\end{tabular}          & \begin{tabular}[c]{@{}c@{}}2\\ 1\end{tabular} & \begin{tabular}[c]{@{}c@{}}2\\ 0\end{tabular} & \begin{tabular}[c]{@{}c@{}}4\\ 2\end{tabular}   & \begin{tabular}[c]{@{}c@{}}256\\ 256\\ 256\end{tabular}          & \begin{tabular}[c]{@{}c@{}}4\\ 1\\ 1\end{tabular}      & \begin{tabular}[c]{@{}c@{}}2\\ 1\\ 1\end{tabular} & \begin{tabular}[c]{@{}c@{}}2\\ 0\\ 0\end{tabular} & \begin{tabular}[c]{@{}c@{}}4\\ 2\\ 2\end{tabular}    & \begin{tabular}[c]{@{}c@{}}512\\ 512\\ 512\end{tabular}           & \begin{tabular}[c]{@{}c@{}}3\\ 1\\ 2\end{tabular}      & \begin{tabular}[c]{@{}c@{}}2\\ 1\\ 2\end{tabular}     & \begin{tabular}[c]{@{}c@{}}1\\ 0\\ 0\end{tabular}     & \begin{tabular}[c]{@{}c@{}}16\\ 4\\ 16\end{tabular}        & \begin{tabular}[c]{@{}c@{}}1024\\ 1024\\ 1024\end{tabular}        & \begin{tabular}[c]{@{}c@{}}3\\ 1\\ 2\end{tabular}      & \begin{tabular}[c]{@{}c@{}}2\\ 1\\ 2\end{tabular}     & \begin{tabular}[c]{@{}c@{}}1\\ 0\\ 0\end{tabular}     & \begin{tabular}[c]{@{}c@{}}32\\ 8\\ 32\end{tabular}        \\ \hline
\begin{tabular}[c]{@{}c@{}}512\\ 512\end{tabular}              & \begin{tabular}[c]{@{}c@{}}3\\ 1\end{tabular}          & \begin{tabular}[c]{@{}c@{}}1\\ 1\end{tabular} & \begin{tabular}[c]{@{}c@{}}1\\ 0\end{tabular} & \begin{tabular}[c]{@{}c@{}}32\\ 4\end{tabular}  & \begin{tabular}[c]{@{}c@{}}512\\ 512\\ 512\end{tabular}          & \begin{tabular}[c]{@{}c@{}}3\\ 1\\ 1\end{tabular}      & \begin{tabular}[c]{@{}c@{}}1\\ 1\\ 1\end{tabular} & \begin{tabular}[c]{@{}c@{}}1\\ 0\\ 0\end{tabular} & \begin{tabular}[c]{@{}c@{}}32\\ 4\\ 4\end{tabular}   & \begin{tabular}[c]{@{}c@{}}1024\\ 1024\\ 1024\end{tabular}        & \begin{tabular}[c]{@{}c@{}}3\\ 1\\ 2\end{tabular}      & \begin{tabular}[c]{@{}c@{}}1\\ 1\\ 2\end{tabular}     & \begin{tabular}[c]{@{}c@{}}1\\ 0\\ 0\end{tabular}     & \begin{tabular}[c]{@{}c@{}}64\\ 8\\ 32\end{tabular}        & \begin{tabular}[c]{@{}c@{}}2048\\ 2048\\ 2048\end{tabular}        & \begin{tabular}[c]{@{}c@{}}3\\ 1\\ 2\end{tabular}      & \begin{tabular}[c]{@{}c@{}}1\\ 1\\ 2\end{tabular}     & \begin{tabular}[c]{@{}c@{}}1\\ 0\\ 0\end{tabular}     & \begin{tabular}[c]{@{}c@{}}128\\ 16\\ 64\end{tabular}      \\ \hline
\begin{tabular}[c]{@{}c@{}}1024\\ 1024\end{tabular}            & \begin{tabular}[c]{@{}c@{}}4\\ 1\end{tabular}          & \begin{tabular}[c]{@{}c@{}}4\\ 1\end{tabular} & \begin{tabular}[c]{@{}c@{}}0\\ 0\end{tabular} & \begin{tabular}[c]{@{}c@{}}64\\ 8\end{tabular}  & \begin{tabular}[c]{@{}c@{}}1024\\ 1024\\ 1024\end{tabular}       & \begin{tabular}[c]{@{}c@{}}4\\ 1\\ 1\end{tabular}      & \begin{tabular}[c]{@{}c@{}}3\\ 1\\ 1\end{tabular} & \begin{tabular}[c]{@{}c@{}}0\\ 0\\ 0\end{tabular} & \begin{tabular}[c]{@{}c@{}}64\\ 8\\ 8\end{tabular}   & \begin{tabular}[c]{@{}c@{}}2048\\ 2048\\ 2048\end{tabular}        & \begin{tabular}[c]{@{}c@{}}3\\ 1\\ 2\end{tabular}      & \begin{tabular}[c]{@{}c@{}}1\\ 1\\ 2\end{tabular}     & \begin{tabular}[c]{@{}c@{}}1\\ 0\\ 0\end{tabular}     & \begin{tabular}[c]{@{}c@{}}128\\ 16\\ 64\end{tabular}      & \begin{tabular}[c]{@{}c@{}}4096\\ 4096\\ 4096\end{tabular}        & \begin{tabular}[c]{@{}c@{}}3\\ 1\\ 2\end{tabular}      & \begin{tabular}[c]{@{}c@{}}1\\ 1\\ 2\end{tabular}     & \begin{tabular}[c]{@{}c@{}}1\\ 0\\ 0\end{tabular}     & \begin{tabular}[c]{@{}c@{}}256\\ 32\\ 128\end{tabular}     \\ \hline
\begin{tabular}[c]{@{}c@{}}2048\\ 2048\end{tabular}            & \begin{tabular}[c]{@{}c@{}}2\\ 1\end{tabular}          & \begin{tabular}[c]{@{}c@{}}1\\ 1\end{tabular} & \begin{tabular}[c]{@{}c@{}}0\\ 0\end{tabular} & \begin{tabular}[c]{@{}c@{}}32\\ 16\end{tabular} & \begin{tabular}[c]{@{}c@{}}2048\\ 4096\\ 4096\end{tabular}       & \begin{tabular}[c]{@{}c@{}}2\\ 1\\ 1\end{tabular}      & \begin{tabular}[c]{@{}c@{}}1\\ 1\\ 1\end{tabular} & \begin{tabular}[c]{@{}c@{}}0\\ 0\\ 0\end{tabular} & \begin{tabular}[c]{@{}c@{}}32\\ 16\\ 32\end{tabular} & \begin{tabular}[c]{@{}c@{}}4096\\ 4096\\ 4096\\ 4096\end{tabular} & \begin{tabular}[c]{@{}c@{}}2\\ 1\\ 1\\ 1\end{tabular}  & \begin{tabular}[c]{@{}c@{}}1\\ 1\\ 1\\ 1\end{tabular} & \begin{tabular}[c]{@{}c@{}}0\\ 0\\ 0\\ 0\end{tabular} & \begin{tabular}[c]{@{}c@{}}128\\ 32\\ 32\\ 32\end{tabular} & \begin{tabular}[c]{@{}c@{}}8192\\ 8192\\ 8192\\ 8192\end{tabular} & \begin{tabular}[c]{@{}c@{}}2\\ 1\\ 1\\ 1\end{tabular}  & \begin{tabular}[c]{@{}c@{}}1\\ 1\\ 1\\ 1\end{tabular} & \begin{tabular}[c]{@{}c@{}}0\\ 0\\ 0\\ 0\end{tabular} & \begin{tabular}[c]{@{}c@{}}256\\ 64\\ 64\\ 64\end{tabular} \\ \hline
4096                                                           & 1                                                      & 1                                             & 0                                             & 16                                              & 8192                                                             & 1                                                      & 1                                                 & 0                                                 & 32                                                   & 8192                                                              & 1                                                      & 1                                                     & 0                                                     & 32                                                         & 16384                                                             & 1                                                      & 1                                                     & 0                                                     & 64                                                         \\ \hline
\end{tabular}
}
\caption{\textbf{Four architectures to study the effects of increasing capacity on performance.} As the student architectures get wider and deeper, more than one TrueNorth chips are tiled to provide more capacity (explained in detail in the hardware implementation section). Here we show architectures used in our experiments deployed on 1,2,4 and 8 TrueNorth chips. Each row represents a layer in the network from the first layer to the last layer going from top to bottom in the table. All network layers are convolutional layers with parameter settings as shown. The structure of the networks used is similar to \cite{Esser11441}.}
\end{table*} 
\subsection{Policy Distillation}
Policy distillation, introduced by Rusu et al. in \cite{rusu2015policy} focuses on transferring knowledge from a learned policy (teacher) to a new network (student) as a supervised learning problem. In this work, the student network, initialized randomly, is trained to mimic the teacher by learning to match the teacher's actions on states sampled from the teacher's trajectories. The teachers are trained using DDQN on the ATARI game suite and provide a dataset $D^T = \{(s_i, \textbf{q}_i)\}_{i=0}^N$ consisting of input observations $s_i$ and corresponding output vector $\textbf{q}_i$.

The authors experiment with three different types of training targets corresponding to three distinct loss functions. The first of these corresponds to training the student to regress to the teacher's learned Q values directly employing a mean squared error (MSE) loss function. 
\begin{equation}
    L_{MSE} (D^T, \theta_{S}) = -\sum_{i=1}^{|D|} ||\textbf{q}_i^T - \textbf{q}_i^S ||_2^2
\end{equation}

The second approach employs a negative log likelihood loss (NLL) and the training target for the student is the teacher's action as a one-hot target ($a_{i,\textrm{best}} = $argmax$(q_i)$).

\begin{equation}
    L_{NLL} (D^T, \theta_{S}) = -\sum_{i=1}^{|D|} \textrm{log} P (a_i = a_{i,\textrm{best}} | x_i, \theta_s)
\end{equation}

The third, and most successful approach is the \textit{distillation} approach proposed by Hinton et al (2014), which uses the Kullback-Leibler (KL) divergence with temperature \textbf{$\tau$}

\begin{dmath}
    L_{KL} (D^T, \theta_{S}) = \sum_{i=1}^{|D|} \textrm{softmax}(\frac{\textbf{q}_i^T}{\tau}) \ln \frac{\textrm{softmax}(\frac{\textbf{q}_i^T}{\tau})}{\textrm{softmax}(\textbf{q}_i^S)}
\end{dmath}

We follow the approach and methodology of the last paper in order to compare results from low precision students against published full precision policy distillation results.

\section{Experiments}
Following closely the experimental setup of \cite{rusu2015policy}, we use DDQN to train a teacher network, and use knowledge distillation on the data generated by the teacher for training the student network. In our setting however, unlike the original work, the goal of training is to enable a low precision student network to distill the policy from a full precision teacher network. It is therefore possible to leverage proven low-precision network training algorithms which produce networks which closely match their high precision counterpart, allowing efficient deployment on energy efficient hardware. This approach is sound since the policy is a deterministic function of the Q function. The optimal action for any state, $s$, the policy is 
\begin{equation}
    \pi(s) = \textrm{argmax}_a Q(s,a)
\end{equation}

\begin{table*}[ht]
 \centering
 \begin{tabular}{|c|c|c|c|c|}
 \hline
 \textbf{Temperature}    & \textbf{0.05}     & \textbf{0.01}     & \textbf{0.005} & \textbf{0.001} \\ \hline
 \textbf{Breakout}       & 0\%*              & \textbf{95.80\%}  & 60.58\%        & 31.57\%        \\ \hline
 \textbf{Beamrider}      & 26.46\%           & \textbf{100.14\%} & 94.16\%        & 61.77\%        \\ \hline
 \textbf{Q*bert}         & \textbf{104.99\%} & 96.42\%           & 97.2\%         & 89.75\%        \\ \hline
 \textbf{Space Invaders} & \textbf{80.87\%}  & 80.44\%           & 73.76\%        & 66.88\%        \\ \hline
 \end{tabular}
 \caption{\textbf{Optimal temperature is the same for low precision policy distillation as \cite{rusu2015policy}.} Score of the low precision student network are shown as a percentage of the high precision teacher for each experiment.  $\tau = 0.01$ gave the best overall scores on the four games.  The highest overall score for each game is shown in bold.}
\end{table*}

\begin{figure*}[ht]
  \includegraphics[width=1\textwidth, ,height=0.4\textwidth]{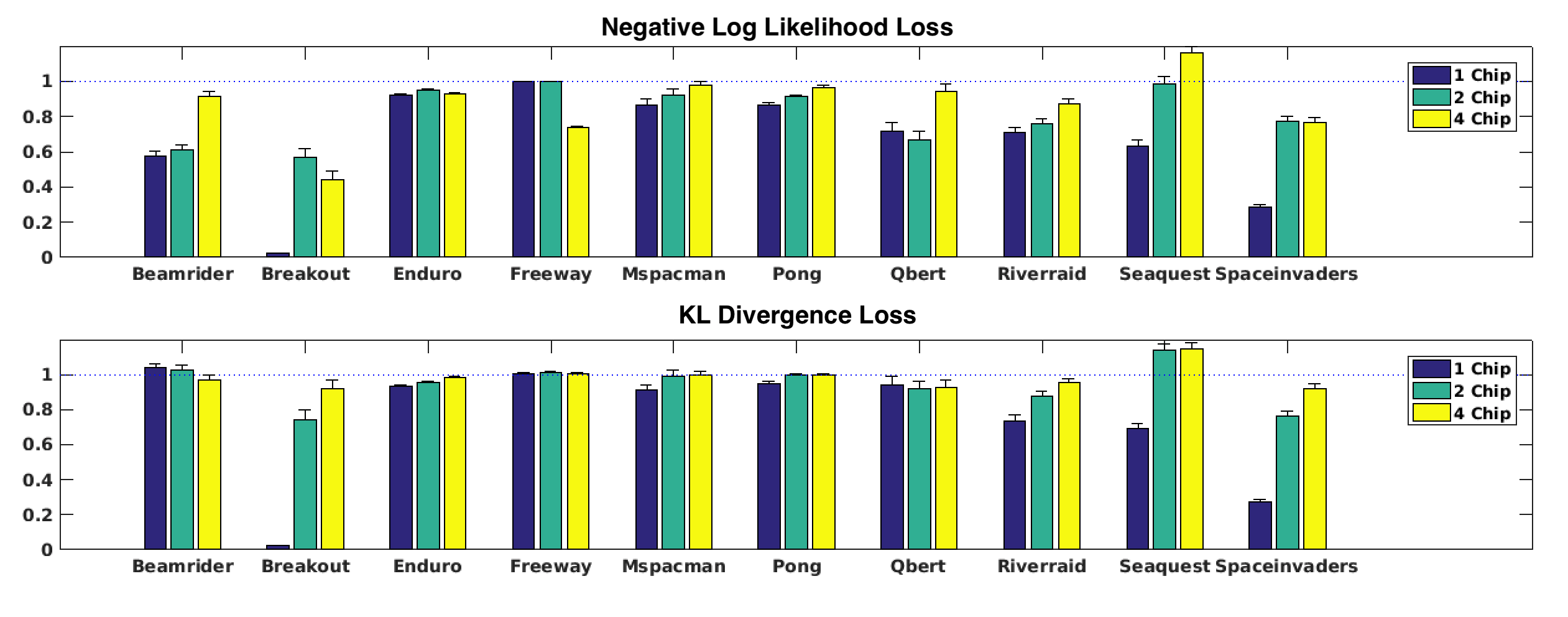}
  \caption{\textbf{Low precision student networks trained using KL Divergence loss function meet or exceed teacher performance for most games.} Policy distillation results for online softmax and KL Divergence student training (mean score for 100 iterations of each game, normalized against the relevant teacher scores).}
  \label{fig:dist_alternate}
\end{figure*}

Therefore, if the teacher network accurately approximates the Q function, then an accurate policy can be derived from it for any state, providing the labels to train the student network. Furthermore, by using policy distillation rather than training a low-precision DQN network directly, the final network is likely to be considerably smaller as well as faster to train. The use of low temperature in the KL divergence loss function makes the labels (Q values) sharper and hence makes it easier for the student to learn the policy of the teacher, without the need for regressing a value function directly from the environment's reward signal. We conjecture that this is one strong reason why we were able to achieve good results with low our low precision algorithms.

We demonstrate our results on single and multi-task policy distillation. For the single task setting, we use a separately trained DDQN for each game as the full precision teacher network with the same parameter settings as \cite{van2016deep}. Specifically, the input frame from the ATARI simulator is converted to grayscale and re-scaled to $84 \times 84$. For every frame, the preceding 3 frames are also given as input, making the input tensor of size $84 \times 84 \times 4$. This is convolved with a convolution layer containing 32 filters of kernel size 8 and stride 4, followed by 64 filters of kernel size 4 and stride 2 in the second convolutional layer, and a final convolutional layer containing 64 filters of of kernel size 3 and stride 1. This is followed by a fully-connected hidden layer of 512 units. The outputs of all layers are passed through Rectifier Linear Units (ReLU) before passing to the next layer. The output of the fully connected layer is projected to the Q valued outputs through a linear layer. Like the original paper, we also trained the network with RMSProp (with momentum parameter 0.95), with the discount set to $\gamma$= 0.99, and the learning rate to $\alpha$ = 0.00025. The number of steps between target network updates was $\tau$ = 10000. We trained the teachers for 50 million frames, using a replay memory buffer of 1 million $(s,a,r,s')$ tuples, and a batch size of 32. We used an $\epsilon$-greedy policy for exploration with the $\epsilon$ decreasing linearly from 1 to 0.1 over the first one million steps.
\begin{figure*}[ht]
  \centering
  \includegraphics[width=0.7\textwidth, ,height=0.33\textwidth]{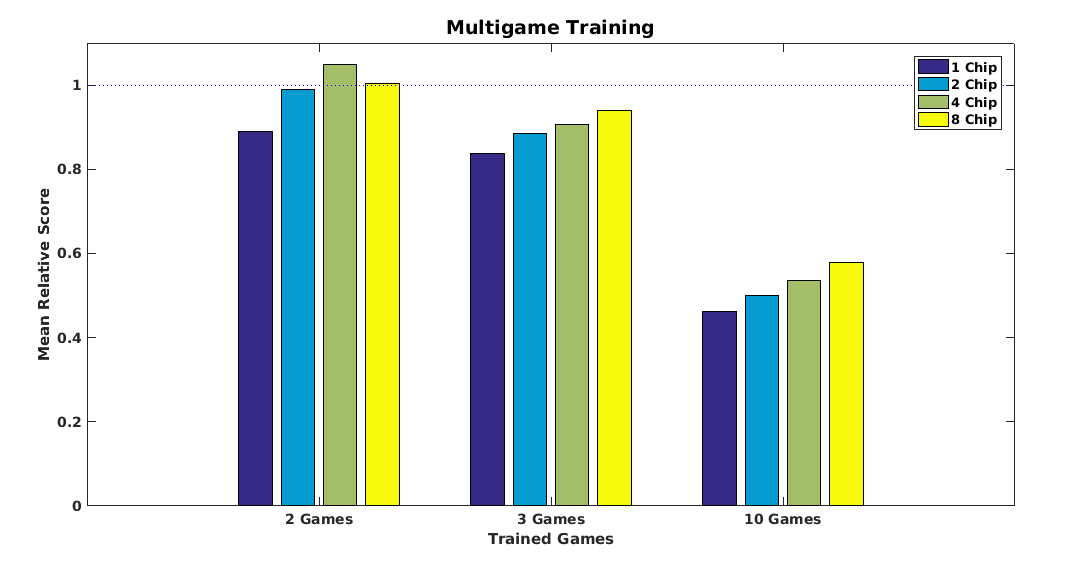}
  \caption{\textbf{Larger student network with sufficient capacity learns to play two games just as well as one.} Policy distillation results for students trained against 2, 3 or 10 ATARI games.  Students were trained against Pong and Qbert;  Pong and Space Invaders (averaged results shown in the 2 game plot above); Pong, Qbert and Space Invaders; Pong, Qbert, and Freeway (averaged results shown in the 3 game plot above); or against all 10 ATARI games shown in Figure \ref{fig:dist_alternate}.  Students were tested against only those games for which they were trained, and their performance is normalized against an identical student trained against only the target game.  For each model variation (1 - 8 chips), normalized results for all tested games were averaged for models trained against 2 games, for those trained against 3 games, and for those trained against 10 games.} 
  \label{fig:multit}
\end{figure*}

For the student, we use a similar architecture with more units in each layer and additional convolution layers to compensate for the hardware constraints. We use four different architectures of different depth and width in each layer for our experiments, the details of which are shown in Table 1. In the student training procedure, the teacher receives frames from the environment and selects actions in an $\epsilon$ greedy manner with $\epsilon$ = 0.05. The Q value outputs of the teacher and the input frames are stored in a buffer of size 1 million. In a training step, the student is trained on a batch of 32 such input-output tuples sampled randomly from the expert buffer. The input to the network is an 8 bit grayscale image, re-scaled to $84 \times 84$. We use a learning rate $\alpha$ of 20, and train the student using SGD (with momentum parameter 0.9). We always use batch normalization with constrained student networks, and since the low precision weights act as a regularizer, we don't decay the weights. The EEDN training framework introduces a spike sparsity parameter $\eta$, which controls the weight of a sparsity inducing penalty added to the loss function given by $\frac{\eta}{2}\sum \bar{y}^2 $, where $\bar{y}$ is the average feature activation and the summation is over all the features of a network. We use $\eta$ = 0.0001 in our experiments. This penalty term serves both as a regularizer and induces sparsity in the outputs of neurons, reducing spike traffic during deployment and therefore reducing the total energy consumption. We use uniform initialization for the weights. For a detailed description of the EEDN training methodology and the associated parameter settings, we refer the reader to the original paper \cite{Esser11441}. We experimented with two different loss functions in this supervised learning setting: 
\begin{itemize}
    \item NLL loss function: The student was trained using the Q values generated by a trained teacher as it played for 6 million frames of a single Atari game, using the NLL loss function.
    \item KL Divergence loss function: Same game play setting as described above; the student was trained against the direct outputs of the teacher, but the KL-divergence loss function introduced in \cite{hinton2015distilling} was used.
\end{itemize}

For our experiments, we used the same subset of ten games from the ATARI suite as in \cite{rusu2015policy} in order to allow for direct comparability of the results. We use a subset of four (similar to \cite{rusu2015policy}) out of the ten games to do a hyper-parameter search for the temperature $\tau$ in $\{0.05, 0.01, 0.005, 0.001\}$. Separate students with the same network architecture and different temperatures were trained for 1.5 million batches (we ran experiments for 6 million iterations and the student was trained on every fourth frame). Consistent with the original work, we found that KL divergence loss performs better than NLL and that $\tau = 0.01$ works best for distillation to low precision student networks, as shown in Table 2.

\begin{table}[ht]
\begin{tabular}{ccc}
                                     & \textbf{NLL}                 & \textbf{KL-Divergence}                \\ \cline{2-3} 
\multicolumn{1}{c|}{\textbf{1 chip}} & \multicolumn{1}{c|}{69.31\%} & \multicolumn{1}{c|}{\textbf{76.71\%}} \\ \cline{2-3} 
\multicolumn{1}{c|}{\textbf{2 chip}} & \multicolumn{1}{c|}{81.45\%} & \multicolumn{1}{c|}{\textbf{87.15\%}} \\ \cline{2-3} 
\multicolumn{1}{c|}{\textbf{4 chip}} & \multicolumn{1}{c|}{94.28\%} & \multicolumn{1}{c|}{\textbf{98.24\%}} \\ \cline{2-3} 
\end{tabular}
\caption{\textbf{Overall, students get within 2\% of teacher scores when trained with KL-Divergence.} Mean scores across all 10 games as a percentage of teacher network scores for networks of different sizes (Table 1) for the two loss functions. Bold entries indicate highest scores for each network size.}
\end{table}

Figure \ref{fig:dist_alternate} shows the results of training single game distillation students for the 10 Atari games normalized against the relevant teacher scores, using the methods described in the training procedure in \cite{esser2015backpropagation} with KL-divergence loss \cite{rusu2015policy} and NLL loss for 1,2 and 4 chip models. Ideally, all students would match their teachers and hit the dashed line at 1. For most games, network size has a direct impact on performance. This is in line with the intuition that the performance is compromised due to the limited capacity owing to the hardware constraints and increasing the capacity helps overcome this effect. The mean across all 10 games for different networks sizes is given in Table 3. First, KL-divergence loss produces better students than NLL for all networks sizes.  Second, The performance of the low-precision student approaches that of the high precision teacher network as the network size increases allowing a trade-off to be made between performance and implementation cost.

As was described in  \cite{rusu2015policy}, we also evaluate multi-game policy distillation. Similar to the full precision case in the original work, we are interested in evaluating whether the low precision students are also expressive enough to be able to learn combined teacher policies from multiple teachers and effectively, is it possible to achieve good performance on multiple games from a single student trained simultaneously from multiple teachers. In this case, training was done offline. Separate teacher models were trained against a single ATARI game, and the argmax of the outputs of the teacher model were saved for 900000 frames as one hot labels. The outputs for 2, 3, or 10 games were merged and randomized, and the student models were trained against the merged training data as before. Effectively, the merged input-output combinations from the teachers serve as the training dataset for supervised training of a single student. Figure \ref{fig:multit} shows the results of these experiments for various combinations of games, for 1, 2, 4 and 8 chip models (more chips correspond to larger networks and higher capacity, see Table 1).  Consistent with the results described in \cite{rusu2015policy}, student models are fully capable of learning 2, 3, or even 10 Atari games, although with reduced performance (when compared to identical models trained against only one game.)   This performance reduction is less pronounced for larger networks, where the student network has more capacity, but persists even in 4 and 8 chip models trained against 3 or more games.  

\section{Hardware Implementation}

The models described above, once trained offline, can be mapped to the TrueNorth chip using ``corelets", as described in \cite{Esser11441} and \cite{amir2013cognitive}. TrueNorth has been integrated onto several different platforms and systems \cite{Sawada:2016:TEB:3014904.3014920}, examples of which are shown in Figure 1. The Neurosynaptic System 1 million neuron evaluation platform (NS1e) is a development platform which contains a single TrueNorth chip alongside a Xilinx Zynq Z-7020 FPGA. Contained within the FPGA are two ARM Cortex-A9 cores which can be used to develop and run applications that take advantage of TrueNorth for its inference capabilities. Through tiling, the TrueNorth chips can also be directly connected to one another via it’s native chip-to-chip asynchronous communication interfaces. With this, we have created a platform which natively tiles 16 TrueNorth chips, the NS16e (Neurosynaptic System 16 million neuron evaluation platform), and provides a fabric which is capable of executing neural networks 16 times larger than that available on the NS1e. Finally, each platform can be scaled-out via packet switch networks and programmed to act as independent nodes or larger capacity neural networks with software orchestrating the communication. Internally, we have deployed a system with an aggregate capacity of 80 million neurons and 20 billion synapses constructed using 80 NS1e boards interconnected via a packet switched network. 

The Arcade Learning Environment (ALE) was augmented for use with all of these systems by providing hooks into the TrueNorth run-time. Prior to transferring input to TrueNorth, the game state needs to be first converted into independent single-bit features \cite{Esser11441}, a process called transduction. Similarly, on the output, the received spikes need to be translated into a class probability corresponding to the action to be taken. In our implementation, we have ported the entire ALE game-engine (Stella) and corresponding software modifications so they are completely run on the ARMs while using TrueNorth to perform inference. During execution, the system is able to maintain a frame-rate of 30 frames-per-second (fps).

\section{Discussion}
To the best of our knowledge, this work is the first of its kind to attempt to adapt reinforcement learning algorithms to specialized low precision hardware in order to be deployed in real world applications, which necessitate energy efficiency and real time inference while ensuring good performance. The results of our experiments with double deep Q networks and policy distillation indicate that low-precision policy distillation is a viable approach to overcome the challenges associated with training extremely reduced precision networks to perform reinforcement learning tasks and hence provides a strong baseline to compare future work targeted at closing the gap between algorithmic advances and real world deployment using highly optimized hardware. This is an important challenge in the current research landscape.

By sidestepping the challenges associated with traditional reinforcement learning, this approach overcomes several issues like credit assignment, noisy gradients and long training duration, and successfully demonstrates mapping of reinforcement learning policies for sequential decision making to energy efficient hardware. Breaking down the task into training full precision teachers using any standard reinforcement learning algorithm off chip, followed by training constrained student networks that mimic the teachers for on chip deployment, also provides the additional advantage of an algorithm agnostic methodology that is shown in this work to work well in practice, and overcomes the challenge of adapting every new reinforcement learning algorithm for the reduced precision space in order to be deployed in a real world application while remaining extremely low power and real time. 

This work can be viewed as a first attempt to adapt value approximation algorithms to low precision networks. Although we are not the first to apply distillation to low precision classifier training \cite{mishra2017apprentice}, this work is the first to demonstrate low precision policy distillation.The training methodology demonstrated here allows for fast training of constrained networks, without compromising on performance during real time deployment. We suspect that many practitioners will find low precision policy distillation useful given that it can be applied to a wide range of emerging value-based algorithms which may be difficult to train directly in the low-precision space.
\subsubsection*{Acknowledgments}
This material is based upon work supported by the Air Force Research Laboratory under Contract No. FA8750-15-C-0121 and FA8750-17-C-0097. 
This report was prepared as an account of work sponsored by an agency of the United States government. Neither the United States Government nor any agency thereof, nor any of their employees, make any warranty, express or implied, or assumes any legal liability or responsibility for the accuracy, completeness, or usefulness of any information, apparatus, product, or process disclosed, or represents that its use would not infringe privately owned rights.
\small
\bibliography{ref.bib}
\bibliographystyle{aaai}
\end{document}